# Single Image Haze Removal using a Generative Adversarial Network

Bharath Raj N.
Electronics and Communication Engineering,
Sri Sivasubramaniya Nadar College of Engineering,
Chennai, India.
bharath15017@ece.ssn.edu.in

Venketeswaran N.
Electronics and Communication Engineering,
Sri Sivasubramaniya Nadar College of Engineering,
Chennai, India.
venketeswarann@ssn.edu.in

*Abstract*—Traditional methods to remove haze from images rely on estimating a transmission map. When dealing with single images, this becomes an ill-posed problem due to the lack of depth information. In this paper, we propose an end-to-end learning based approach which uses a modified conditional Generative Adversarial Network to directly remove haze from an image. We employ the usage of the Tiramisu model in place of the classic U-Net model as the generator owing to its higher parameter efficiency and performance. Moreover, a patch based discriminator was used to reduce artefacts in the output. To further improve the perceptual quality of the output, a hybrid weighted loss function was designed and used to train the model. Experiments on synthetic and real world hazy images demonstrates that our model performs competitively with the state of the art methods.

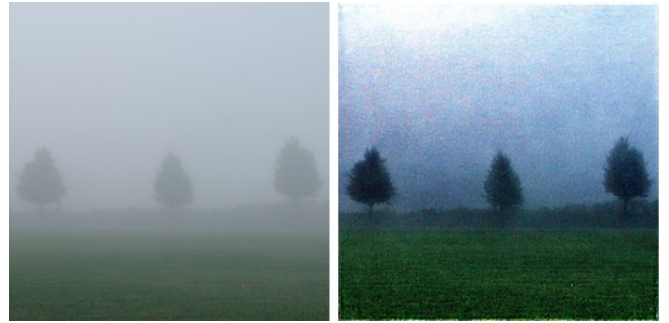

Fig. 1. Sample dehazing results using the proposed method. Left: Input hazy image, Right: Output dehazed image.

## I. INTRODUCTION

Haze is a natural phenomenon caused due to particles in the atmosphere absorbing and scattering light. Due to this, in an hazy atmosphere, light emanating from distant sources are often scattered, and the observer perceives a reduction in contrast. Moreover, images with haze have reduced visibility and color fidelity. Because of this, haze removal is highly sought after where loss in contrast and visibility is of prime importance, such as autonomous vehicles. It also highly desired in photography and other computer vision applications.

$$I(x) = J(x)T(x) + A(x)(1 - T(x)) \quad (1)$$

A hazy I(x) image can be approximately parameterized by equation (1). Here, x is the pixel coordinates in the image. J(x) is its haze-free counterpart and T(x) is the transmission map [1]. The term A(x) is the global atmospheric light. When atmospheric light is homogeneous, the transmission map is given by:

$$T(x) = e^{-\beta d(x)} \quad (2)$$

Here, $\beta$ refers to the attenuation coefficient and d(x) refers to the scene depth. By estimating the transmission map, we can recover the original haze-free image. Accurate, straightforward recovery of the transmission map requires depth information. Several dehazing algorithms have obtained depth information by considering multiple images of the same scene, 3D modelling or images in different polarizations [2, 3].

Single Image dehazing is an under constrained problem. Classical single image dehazing algorithms are dependent on the assumption of a good prior. Accordingly, several methods have been proposed for dehazing using priors [4, 5]. But, the performance of these algorithms depend upon the accuracy of the priors, and they often fail to generalize for all types of scenes.

Another class of single image dehazing algorithms are based on machine learning algorithms [6, 7, 8]. Some try to estimate the transmission map or haze relevant features directly, without assuming a prior, and then solve equation (1) to get the haze-free image J(x). Since these methods do not require a manually calculated prior, they often generalize to a broader number of scenes. Often the euclidean loss function is used to optimize the algorithms to estimate the transmission map and then the haze-free image is manually calculated using equation (1). While this provides decent results, the created haze-free image often has artefacts.

To overcome these limitations, better learning algorithms and loss functions can be designed. Generative Adversarial Networks (GANs) [9] and conditional GANs [10] have shown remarkable promise in several image translation and reconstruction tasks [11, 12]. Moreover, loss functions such as perceptual loss [13] can improve the visual quality of the image. Keeping the above points in mind, we design an end-to-end trainable conditional GAN that can remove haze provided only a single image of the scene. We elaborate our contributions in the below sub-section:

### A. Contributions

- We introduce the usage of a conditional GAN (cGAN) to directly remove haze given a single image of the scene.
- Typically, the generator of a conditional GAN uses the U-Net [14] architecture. We replace it with the 56-Layer Tiramisu model. The latter is parameter efficient, and has state of the art performance in semantic segmentation [15].
- We use the Patch Discriminator [10] in our conditional GAN to reduce artefacts.
- A weighted loss function is designed to include the effect of L1 Loss and Perceptual Loss in addition to the standard conditional GAN generator loss to motivate the model to generate visually appealing outputs.

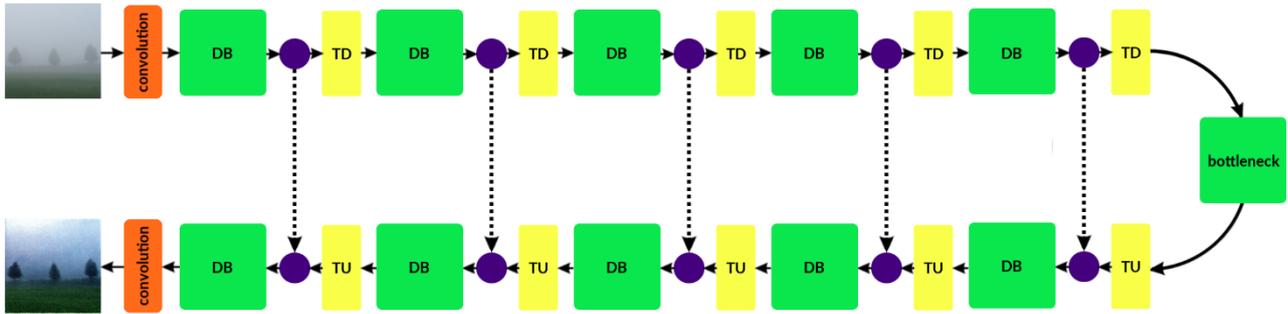

**Fig. 2.** Generator of the proposed model. It consists of Dense Blocks (DB), Transition Down layers (TD), Transition Up layers (TU) and a Bottleneck layer. Additionally, it contains an input convolution and an output convolution. Dotted lines imply a concatenation operation.

## II. RELATED WORK

Classic haze removal requires us to provide the depth information of the image, so that the transmission coefficient can be calculated at every pixel. A transmission map is hence obtained, and is used recover the original RGB image by using equation (1). Kopf et. al [2] did this by calculating the depth from multiple images, or a 3D model.

Single image dehazing is under-constrained, as we do not have additional depth information. Several approaches under this domain involve the creation of an image prior, based on certain assumptions. Most of these approaches work locally, on patches of the image, with maybe prone to artefacts. They apply guided filtering to smoothen and reduce such artefacts.

Tan [5] observed that haze-free images must have higher contrast than its hazy counterpart, and attempts to remove haze by maximizing the local contrast of the image patch. He et. al. [4] found that most patches have pixels that have a very low intensity in at least one color channel. In its hazy counterpart, the intensity of the pixel in that channel is mostly contributed by airlight. Using this information, a dark channel prior is calculated, and is used for dehazing as well as estimating the transmission map of the image. The drawback is that, removing haze from regions which naturally have high intensity, such as the sky, is error prone. Improving on that, Meng et. al [16] imposed boundary constraints on the transmission function to make the transmission map prediction more accurate. Berman et. al [17] suggested a Non-local approach to perform dehazing. They observed that colours in a haze-free image can be approximated by a few 100 distinct colours that form clusters in the RGB space. They find that pixels in an hazy image can be modelled as lines in the RGB space (haze lines), and using this, they estimate the transmission map values at every pixel.

All of these methods are based on one or more key assumptions, which exploit haze relevant features. Some of these assumptions do not hold true in all possible cases. A way to circumvent this issue is to use deep learning techniques, and let the algorithm decide the relevant features. DehazeNet is a CNN that outputs a feature map, from which the transmission map is calculated [6]. Going a step further, Ren et. al [7] have calculated the transmission map using a multiscale deep CNN.

Conditional GANs [10] have proved to be immensely effective in several image translation applications such as Super Resolution, De-Raining, and several others [11, 12]. Zhang et. al. [8] used three modules of neural networks, to perform guided dehazing of an image. The first module is a conditional GAN used to estimate the transmission map. The second module estimates the haze relevant features, and is concatenated with the transmission map. Finally, the third module translates this concatenated combination to obtain a haze-free image. While this approach is systematic, usage of three modules makes the learning process computationally expensive and parameter heavy.

## III. PROPOSED MODEL

In this paper, we introduce an end to end trainable, conditional Generative Adversarial Network, which is capable of removing haze without estimating the transmission map explicitly. The components of the network, namely the Generator, Discriminator and the Loss Functions are described in the following subsections.

### A. Generator

We replace the conventional U-Net [14] used as the generator, with the 56 Layer Tiramisu [15]. This enhances information and gradient flow, due to the presence of dense blocks [18]. The model has 5 Dense Blocks on the encoder side, 5 Dense Blocks on the decoder side and one Dense Block as the bottleneck layer.

On the encoder side, each Dense Block (DB) is followed by a Transition Down (TD) layer. The TD layer comprises of batch normalization, convolution, dropout and average pooling operations. Similarly, on the decoder side, each Dense Block is followed by a Transition Up (TU) layer. The TU layer has only a transpose convolution operation. Each DB layer in the encoder and the decoder has 4 composite layers each. The DB layer in the bottleneck has 15 composite layers. Each composite layer is comprised of batch normalization, relu, convolution and dropout operations. The growth rate for each DB layer is fixed at 12.

The spatial dimensions of the images are halved after passing through a TD layer, and doubled after passing through a TU layer.

### B. Discriminator

We use the same patch discriminator network as used in the original conditional GAN paper [10]. This network performs patch-wise comparison of the target and generated images, rather than pixel-wise comparison. This is enabled by passing the images through a CNN, whose receptive field at the output is larger than one pixel (i.e. corresponds to a patch of pixels in the original image). Valid padding is used to control the effective receptive field. We follow the 70x70 patch discriminator, as described in the Pix2Pix GAN paper [10]. We apply pixel-wise comparison at the last set of feature maps. The effective receptive field at the feature map is larger than a pixel, and hence it covers a patch of the images. This removes a good amount of artefacts in the images [10].

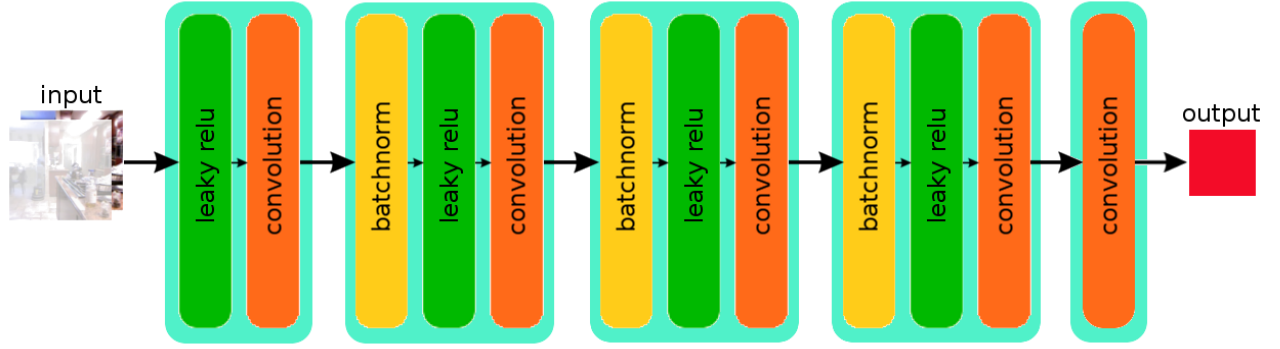

**Fig. 3.** Discriminator of the proposed model. It takes in an input of an hazy image concatenated with the ground truth or the generated image. It outputs a 30x30 matrix, which is used to test if the image is real or fake.

*C. Loss Functions*

We define the standard cGAN objective function to train the generator and discriminator networks as:

$$L_A = \mathbb{E}_{x,y}[\log(D(x,y))] + \mathbb{E}_x[\log(1 - D(x, G(x)))] \quad (3)$$

Here, **D** refers to the discriminator and **G** refers to the generator. The input hazy image is defined as **x**, and its corresponding haze-free counterpart is defined as **y**. While training the cGAN, the generator tries to minimize the objective whereas the discriminator tries to maximize it.

More specifically, given a list of hazy images X=[$x_1$, $x_2$, …, $x_N$] and its corresponding set of haze-free images Y=[$y_1$, $y_2$, …, $y_N$], the standard generator objective is to minimize $L_G$ as stated below:

$$L_G = -MEAN(\log(D(X, G(X)))) \quad (4)$$

Instead of using the standard generator loss directly, we augment it with L1 ($L_{L1}$) and Perceptual ($L_{vgg}$) losses to create a **weighted** generator loss function ($L_T$). This is performed to improve the quality of our generated images. We **minimize $L_T$ directly** to train our generator.

$$L_T = W_{gan} L_G + W_{L1} L_{L1} + W_{vgg} L_{vgg} \quad (5)$$

The discriminator will **maximize** the following objective function $L_D$:

$$L_D = MEAN(\log(D(X,Y)) + \log(1 - D(X, G(X)))) \quad (6)$$

The following subsection will elaborate on the L1 Loss and the Perceptual loss.

*1) L1 Loss*

As stated in [10], using a weighted L1 Loss along with the adversarial loss reduced artefacts in the output image. The L1 Loss between the target image y and the generated image $G(x)$ is calculated as follows:

$$L_{L1} = \mathbb{E}_{x,y}[\|y - G(x)\|_1] \quad (7)$$

The result is multiplied with a scalar weight $W_{L1}$ and is added to the total loss.

*2) Perceptual Loss*

Similar to the following papers [11, 12], we added the feature reconstruction perceptual loss to our total loss. However instead of L2 we use the Mean Squared Error (MSE) scaled by a constant C. The generated and target images are passed through a non-trainable VGG-19 network. The MSE Loss of the outputs of the two images, after passing through the Pool-4 layer, is calculated. The perceptual loss is defined as follows:

$$L_{vgg} = C * MSE(V(G(x)), V(y)) \quad (8)$$

The value of the constant C was empirically set to 1e-5. The function V represents the non-linear CNN transformation which is performed by the VGG network. The result is multiplied with a scalar weight $W_{vgg}$ and is added to the total loss.

IV. EXPERIMENTS

*A. Dataset*

To create a realistic haze dataset, we need depth information. Since it is difficult to find a large, realistic haze/ground-truth image pair dataset, we synthetically created one using the NYU Depth [19] dataset and the Make 3D [20, 21] dataset. The NYU depth dataset contains 1449 indoor scenes along with their corresponding RGB-D pair, and the Make 3D dataset contains images of outdoor scenes along with their depth information. The transmission map (t) for the images were created using 1 minus normalized depth and the atmospheric light factor (A) was set to 1. The image and depth information were scaled to size (256, 256). We then used equation (1) to create hazy images for a given ground-truth image. To simulate different intensities of heavy-haze the transmission map for each image was scaled by a random value sampled from the uniform distribution [0.2, 0.4]. The resultant dataset after combining both datasets had a total of 1776 haze/ground-truth image pairs.

*B. Training Details*

The weight values were empirically chosen as $W_{vgg} = 10$, $W_{gan} = 2$, $W_{L1} = 100$. The learning rate value was fixed at 0.001. We used the Adam Optimizer to perform gradient descent. We perform one update of the generator followed by one update of the discriminator per iteration.

The dataset was split into a training set, validation set and a test set. A set of 1550 images were randomly chosen from the dataset, to create the training dataset. Additionally, these images were flipped horizontally and added to the training dataset, creating a total of 3100 images. The remaining 226 images was split into a validation set of 76 images and a test set of 150 images. Due to the rather small size of the dataset, we used very small validation and test sets. Nevertheless, this was sufficient to assess the generalization capability of the model and pick out the best performing model.

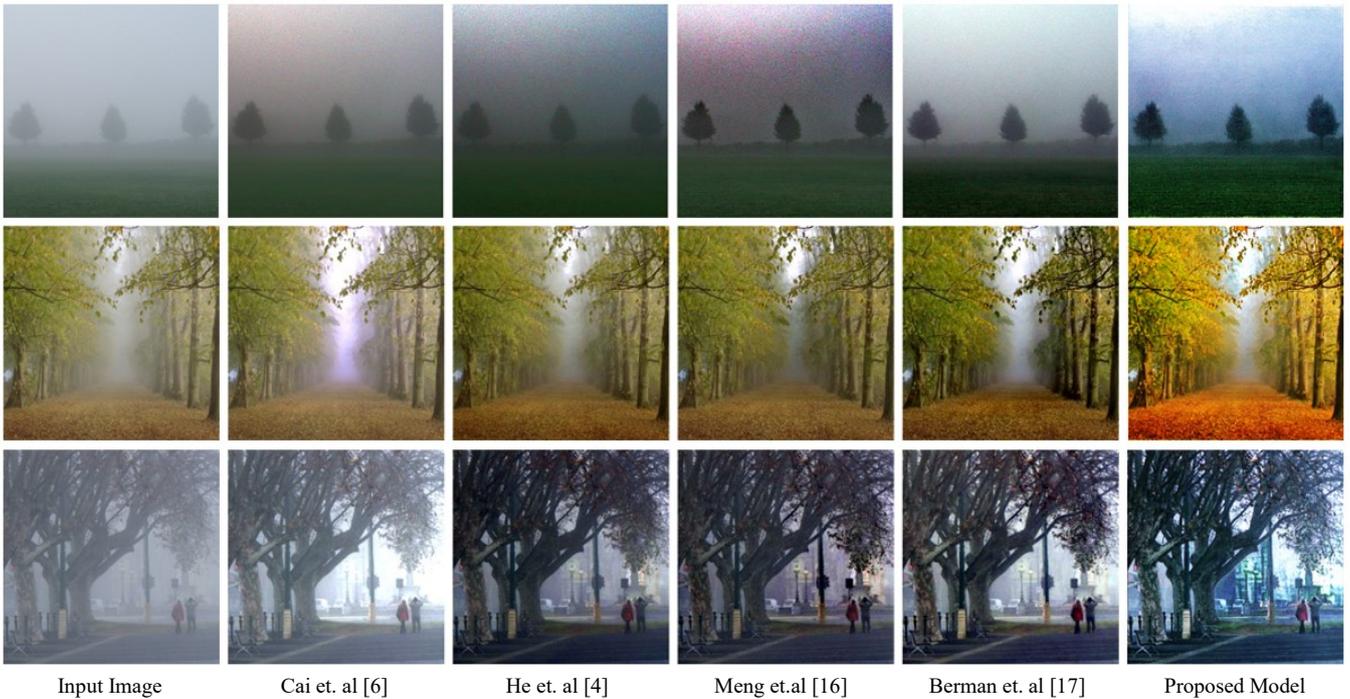

| Input Image | Cai et. al [6] | He et. al [4] | Meng et.al [16] | Berman et. al [17] | Proposed Model |

**Fig. 4.** Visual comparison of dehazing method on real world images.

To further evaluate performance on outdoor images, we used the Synthetic Objective Testing Set (SOTS) subset of the RESIDE [22] dataset. The SOTS subset contained 500 pairs of outdoor images with and without haze. We did not use the indoor images from the SOTS subset for our experiments. Quantitative results are tabulated in Table 1.

*C. Performance Metrics*

A haze removal algorithm's performance can be evaluated on several factors, among them, two of the most frequently used factors are the PSNR and SSIM. Peak Signal to Noise Ratio (PSNR) measures the ability of the algorithm to remove noise from a noisy image. Two identical images will have a PSNR value of infinity.

For measuring haze removal capability, a higher PSNR value indicates better performance. Structural Similarity Index Measure (SSIM), measures how similar two images are. Two identical images will have a SSIM of 1.

On the contrary, there's no guarantee that images with high PSNR will be visually pleasing and images with high SSIM will have a good haze removal quality. Hence, we need a metric that has both properties. We define a metric called score, which is a weighted sum of the PSNR and the SSIM of the image.

$$Score = W_{PSNR} * PSNR + W_{SSIM} * SSIM \qquad (9)$$

The PSNR weight ($W_{PSNR}$) is set to 0.05 and the SSIM weight ($W_{SSIM}$) is set to 1. During the training process, the model is saved whenever it reaches a higher score on the validation set. The saved model with the highest validation score is used for evaluating results on the test set.

*D. Quantitative and Qualitative Evaluation*

In this section, we given a quantitative and a visual comparison of our model with other popular dehazing techniques. We compare the PSNR and SSIM values obtained by these models on our test set. For our quantitative analysis, the following models are selected:

- Dark channel prior haze removal (He et. al, [4])
- Boundary constraint and context regularization (Meng et. al, [16])
- Non local image dehazing (Berman et. al, [17])

Table 1 lists out the PSNR, SSIM and our score values for the quantitative analysis. Clearly, our model performs really well on our synthetic test dataset. The rather lower score of the other models can be attributed to the fact that our synthetic test dataset has very heavy haze, which may not be in sync with the assumptions taken by the algorithms.

| Metrics | Custom Dataset | | | | SOTS | | | |
|---|---|---|---|---|---|---|---|---|
| | *He et al.* | *Meng et al.* | *Berman et al.* | *Proposed Model* | *He et al.* | *Meng et al.* | *Berman et al.* | *Proposed Model* |
| **PSNR** | 13.89 | 14.48 | 12.48 | 20.32 | 16.97 | 15.59 | 18.11 | 18.75 |
| **SSIM** | 0.659 | 0.651 | 0.649 | 0.759 | 0.829 | 0.809 | 0.839 | 0.790 |
| **Score** | 1.354 | 1.375 | 1.274 | 1.775 | 1.678 | 1.588 | 1.744 | 1.728 |

**Table 1.** Results of the quantitative analysis conducted on our synthetic test set (Custom Dataset) as well as the outdoor images from the SOTS dataset. Our synthetic test dataset has 150 images of extremely intensive haze. Our model performs much better in Custom Dataset, as it is trained on similar images. It also achieves competitive performance in the outdoor SOTS dataset.

Moreover, it reinforces the idea that our model performs very well to undo the synthetic transformation that we applied.

To conclusively prove the effectiveness of our model, we evaluated it on some real world hazy images. Moreover, we compare our results with those of the dehazing methods listed in this paper. The original hazy image, along with the dehazed counterparts are illustrated in Fig. 4.

In Fig. 4, from the first row of images, it is evident that our algorithm performs really well in extreme haze conditions. It can be seen that our algorithm is able to extract more amount of intricate details from the input hazy images than the other methods.

## V. Conclusion

This paper presented a conditional generative adversarial network, that directly removes haze given a single image of the scene without explicitly estimating the transmission map. We employed the usage of the Tiramisu model instead of the classic U-Net model as the generator. By using the patch discriminator and L1 loss and perceptual loss, we reduced the presence of artefacts significantly. We provided a quantitative and visual comparison of our model with other popular dehazing algorithms, proving the superior quality of our model. However, the size of the input image is restricted (256x256), but further work can be done to make this flexible. Also, larger Tiramisu models can be experimented with.